\documentclass[11pt]{article}

\usepackage[]{latex/acl}

\usepackage{times}
\usepackage{latexsym}

\usepackage[T1]{fontenc}

\usepackage[utf8]{inputenc}

\usepackage{microtype}

\usepackage{inconsolata}

\usepackage{graphicx}
\graphicspath{ {./img} }
\usepackage{xcolor}
\usepackage{pifont}
\usepackage{csquotes}
\usepackage{amsmath,amssymb}
\usepackage[frozencache,cachedir=_minted-acl_latex]{minted}
\usepackage{colortbl}
\usepackage{booktabs}
\usepackage{amsmath}
\usepackage{algpseudocode}
\usepackage{algorithm}
\usepackage{upgreek}
\usepackage{bbm}
\usepackage{multirow}
\usepackage{subcaption} 
\usepackage{tcolorbox}

\graphicspath{{img/}}
\definecolor{myred}{RGB}{220, 50, 32}
\definecolor{myblue}{RGB}{0, 90, 181}
\algnewcommand\algorithmicforeach{\textbf{for each}}
\algdef{S}[FOR]{ForEach}[1]{\algorithmicforeach\ #1\ \algorithmicdo}
\algnewcommand{\LineComment}[1]{\State \(\triangleright\) #1}

\newcommand{\numcorpora}{ten }
\newcommand{\numkbs}{six }

\newcommand{\numoutperform}{six }
\newcommand{\ours}{BEL\tcbox[
		on line,
		left=1pt,
		right=2pt,
		top=1pt,
		bottom=1pt,
		boxsep=2pt,
		boxrule=2pt,
		colback=black,
		colframe=lightgray]{\textcolor{white}{\textcolor{lightgray}{\textsl{HD}}}}}

\title{\ours: Improving Biomedical Entity Linking with Homonoym Disambiguation}

\author{Samuele Garda \and  Ulf Leser \\
	Humboldt-Universit\"at zu Berlin \\
	\texttt{\{gardasam,leser\}@informatik.hu-berlin.de}
}

\begin{document}
\maketitle

\begin{abstract}
	Biomedical entity linking (BEL) is the task of grounding entity mentions to a knowledge base (KB). A popular approach
	to the task are name-based methods, i.e. those identifying the most appropriate name in the KB for a given mention,
	either via dense retrieval or autoregressive modeling.
	However, as these methods directly return KB names, they cannot cope with homonyms, i.e. different KB entities sharing
	the exact same name. This significantly affects their performance, especially for KBs where homonyms account for a
	large amount of entity mentions (e.g. UMLS and NCBI Gene).
	We therefore present \textbf{BELHD} (\textbf{B}iomedical \textbf{E}ntity \textbf{L}inking with \textbf{H}omonym
	\textbf{D}isambiguation), a new name-based method that copes with this challenge. Specifically, BELHD builds upon the
	BioSyn~\cite{BiomedicalEntiSung2020} model introducing two crucial extensions. First, it performs a preprocessing of
	the KB in which it expands homonyms with an automatically chosen disambiguating string, thus enforcing unique linking
	decisions. Second, we introduce \textit{candidate sharing}, a novel strategy to select candidates for contrastive
	learning that enhances the overall training signal. Experiments with \numcorpora corpora and five entity types show
	that BELHD improves upon state-of-the-art approaches, achieving the best results in \numoutperform out \numcorpora
	corpora with an average improvement of 4.55pp recall@1. Furthermore, the KB preprocessing is orthogonal to the core
	prdiction model and thus can also improve other methods, which we exemplify for GenBioEL \cite{GenerativeBiomYuan2022},
	a generative name-based BEL approach. Code is available at: link added upon publication.
\end{abstract}

\section{Introduction}\label{sec:intro}

\begin{figure*}
	\centering
	\includegraphics[scale=0.20]{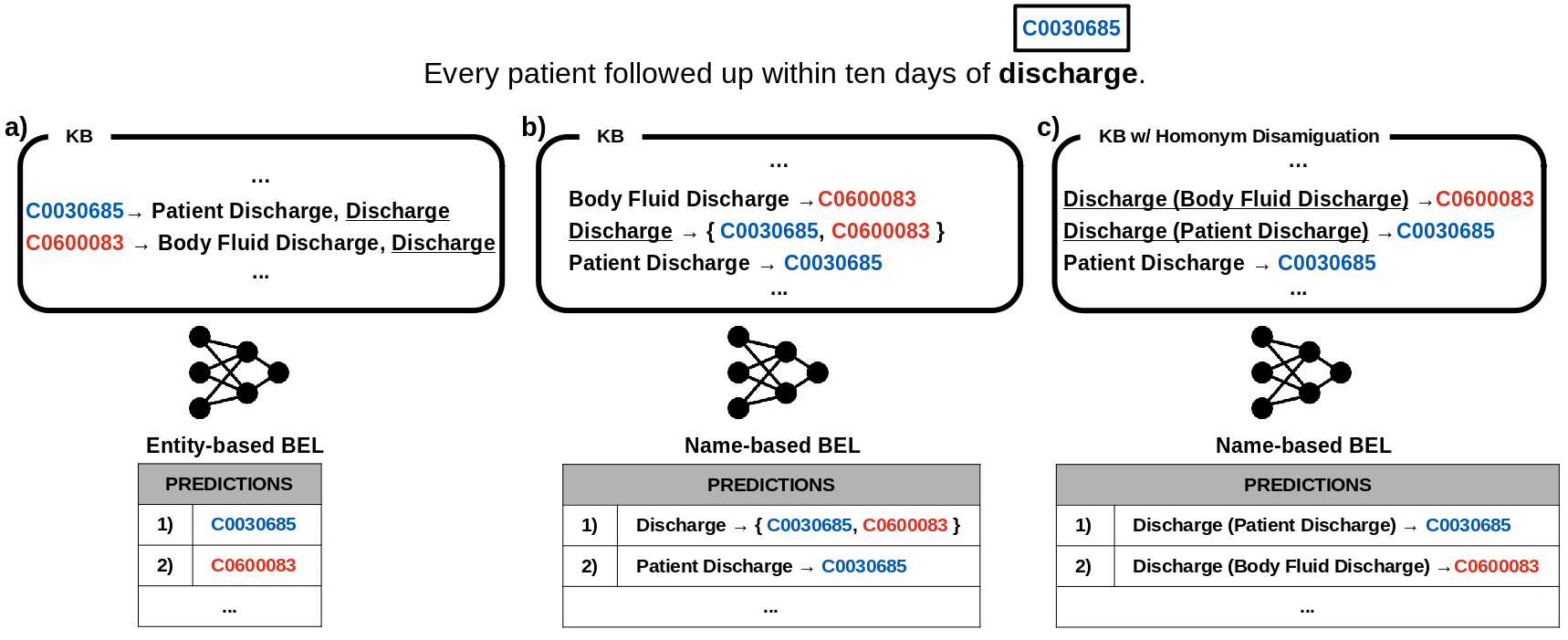}
	\caption{Illustration of entity-based (a) and name-based (b) approaches to biomedical entity linking.
		\underline{Underlined} text highlights the KB homonym (Section \ref{sec:homonyms})
		preventing a unique linking decision (b).
		In (c) we show how in BELHD we address the issue by replacing homonyms their disambiguated version.
		Text in \textcolor{myblue}{blue} and \textcolor{myred}{red} represent the correct and wrong prediction, respectively.}
	\label{fig:name_vs_entity_based}
\end{figure*}

Biomedical entity linking (BEL) is the task of grounding text mentions to a Knowledge Base (KB)
Approaches to BEL can be divided into two categories\footnote{We exclude methods requiring a candidate list, i.e.
	reranking approaches like cross-encoders \cite{ScalableZeroSWuLe2020}.}.
Entity-based methods explicitly construct entity representations usually in form of embeddings for dense retrieval
(\citealp{CrossDomainDaVarma2021,zhang2022knowledge,EntityLinkingAgarwa2022} inter alia). In contrast, name-based
methods directly identify the best matching name in the KB, either via dense retrieval or autoregressive modeling (see
Figure \ref{fig:name_vs_entity_based}).

Though name-based methods have been widely investigated and often outperform other approaches in evaluations
(\citealp{BiomedicalEntiSung2020, SelfAlignmentLiuF2021,GenerativeBiomYuan2022}), they all suffer from a critical flaw,
which is the treatment of homonyms. A homonym is a name in a KB that appears more than once, which results from
multiple KB entities having the same name. As shown in Figure \ref{fig:name_vs_entity_based}, since name-based methods
return the name as a result of linking, they are not capable of resolving such cases~\cite{zhang2022knowledge}.
%
Previous evaluations applied evaluation schemes in which the model is allowed to return multiple KB entities which are
then counted as partly correct.
However, in standard applications BEL is just one step in a complex pipeline in which downstream components typically
cannot properly handle non-unique linking information \cite{BeedsLargeScWang2022}.

We therefore introduce \textbf{BELHD} (\textbf{B}iomedical \textbf{E}ntity \textbf{L}inking with \textbf{H}omonym
\textbf{D}isambiguation), which extends the name-based method BioSyn \cite{BiomedicalEntiSung2020} to properly handle
homonyms. That is, we disambiguate all KB names in a preprocessing step by expanding any instance of a homonym with a
disambiguating string. An example is shown in Figure \ref{fig:name_vs_entity_based} (c), in which \enquote{Discharge}
is a homonym referring the UMLS entities C0030685 and C0600083. We replace both instances with properly expanded names,
i.e. \enquote{Discharge (Patient Discharge)} and \enquote{Discharge (Body Fluid Discharge)}.
Secondly, similar to~\cite{HighlyParallelDeCao2021}, BELHD processes the entire text with \textit{all} of its mentions
at once. This allows us to introduce \textit{candidate sharing}, a novel strategy for contrastive
learning~\cite{ContrastiveRepLeKha2020} in which each mention not only uses its own candidates, but also those
retrieved for other mentions appearing in the same context. As co-occurring mentions are often related, shared
candidates can act as additional hard positive/negative samples enhancing the overall training signal (see Figure
\ref{fig:belbert_training} for an example).

To evaluate our BELHD, we performed extensive experiments to compare its results to entity-based models, arboEL
\cite{EntityLinkingAgarwa2022} and KRISSBERT \cite{zhang2022knowledge}, and name-based ones, BioSyn (without our
extensions) \cite{BiomedicalEntiSung2020} and GenBioEL \cite{GenerativeBiomYuan2022}, on \numcorpora corpora linked to
\numkbs KBs. Overall, we find that BELHD outperforms state-of-the-art approaches in \numoutperform of the \numcorpora
corpora with an average improvement of 4.55pp recall@1. As our approach for homonym disambiguation is independent of
the core model it can also be applied to improve other name-based methods, which we show for the case of GenBioEL.

\section{Related Work}\label{sec:rw}

\noindent \textbf{BEL models} (excluding reranking ones)
can be roughly divided into two categories (Section \ref{sec:intro}).
The first one includes entity-based methods, i.e. those learning a representation for each entity.
\citet{EntityLinkingAgarwa2022} propose arboEL, the current state-of-the-art approach for \texttt{MedMentions},
which concatenates all entity names to form an entity embedding
and constructs k-nearest neighbor graphs over co-referent mention and entity clusters.
Using a pruning algorithm they generate directed minimum spanning trees rooted at entity nodes used for linking.
Though not strictly entity-based,
KRISSBERT \cite{zhang2022knowledge} uses \enquote{entity prototypes} for linking,
which are contextualized mentions of UMLS names (discarding homonyms) retrieved via exact string-matching from PubMed (one of the largest
archive of biomedical literature).
UMLS entities never mentioned in PubMed
are represented by embeddings obtained with their names, semantic hierarchy and descriptions.
The second category consists of name-based models, primarily bi-encoder architectures,
all of them encoding mentions without context.
Notable examples are (i) BioSyn \cite{BiomedicalEntiSung2020}, proposing
a loss function to encourage KB names of the same entity to be closer in the dense space,
(ii) SapBERT \cite{SelfAlignmentLiuF2021}, presenting a pre-training strategy
based on self-supervision shown to improve BioSyn performance and
(ii) follow-up studies replacing BERT \cite{BertPreTrainDevlin2019} with a CNN
to reduce the demanding memory footprint \cite{BertMightBeOLaiT2021,ALightweightNChen2021}.
\citet{GenerativeBiomYuan2022} instead introduce GenBioEL, 
an adaptation for the biomedical domain of GENRE, the autoregressive approach to entity linking first introduced by \citet{decao2021autoregressive}.


\noindent \textbf{Homonyms} are a well known issue in BEL \cite{CrossSpeciesGWeiC2011,DnormDiseaseLeaman2013}.
Recently, \citet{BelbABiomediGarda2023} introduce a comprehensive BEL benchmark
and note that previous studies have focused on only two class of homonyms.
The first one is abbreviations, e.g.\enquote{TS} being either \enquote{Tourette Syndrome} or \enquote{Timothy Syndrome}
while the second one are cross-species genes, e.g. human vs mouse \enquote{$\alpha$2microglobulin}.
State-of-the-art approaches address these instances via specialized tools:
Ab3P \cite{AbbreviationDeSohn2008} for abbreviations and
SR4GN \cite{Sr4gnASpecieWeiC2012} or SpeciesAssignment \cite{AssigningSpeciLuoL2022} for cross-species genes.
These however are non-adaptable solutions crafted for a specific category of homonyms,
leaving many cases unhandled, as the one in the example in Figure \ref{fig:name_vs_entity_based}.



\noindent \textbf{KB augmentation} has been explored in previous work,
though not specific for homonyms.
\citet{procopio-etal-2023-entity} extend Wikipedia entities,
represented by article titles, with their description in WikiData,
e.g. \enquote{Ronaldo} with \enquote{Brazilian football player}.
To improve GENRE's generalization abilities
\citet{OnTheSurprisiSchuma2023} propose to augment Wikipedia entities
with keywords extracted from their descriptions with an unsupervised method.
Both approaches however are solutions specific for Wikipedia,
while ours targets the biomedical domain.
Secondly, in BEL, entity descriptions are seldom available \cite{zhang2022knowledge}.





\section{Background}\label{sec:homonyms}

\noindent \textbf{Task} We formulate BEL as the task of predicting an entity $e \in E$ from a KB
given a document $d$ and a pair of start and end positions $\langle m_{s}, m_{e} \rangle$ indicating a span in $d$
(the entity mention $m$).
In all experiments we use in-KB \cite{GerbilBenchmRoder2018} gold mentions. That is, after the BEL-step, each $m$ is associated to a KB entity.

\noindent \textbf{Biomedical KBs and homonyms}. Each entity $e$ in the KB
is represented with a unique ID associated with a set of names $s \in S$.
E.g. in UMLS the entity C0030685 has the following names: \enquote{Patient Discharge} and \enquote{Discharge}.
As shown in Figure \ref{fig:name_vs_entity_based}, this entails that there are two ways to represent the KB: (a)
by entity or (b) by name. The latter requires defining a mapping $\mathcal{V}_{KB}: S \rightarrow E$,
which for a given $s$ returns its associated entity. The exact same name however can
appear more than once in the KB and thus points to multiple entities. In this case we call the name a \textit{homonym}, \enquote{Discharge} in our example.
Formally, a name $s$ is a homonym if $\vert \mathcal{V}_{KB}(s) \vert > 1$, where $\vert \mathcal{V}_{KB}(s) \vert$ is
the number of entities $s$ maps to.

\noindent \textbf{Impact of homonyms} In case of name-based systems, homonyms fundamentally disrupt
linking, as for mentions linked to them it is impossible to assign a unique KB entity.
Determining to what extent this issue impacts name-based methods in general is however non-trivial.
This is because the number of mentions linked to homonyms depends on the specific model. For instance, for the mention
\enquote{discharge} in Figure \ref{fig:name_vs_entity_based}, it is possible for a name-based system to return
\enquote{Patient Discharge}. However, due to the high similarity between surface forms, we expect most current models
to rank \enquote{Discharge} higher. We can obtain an \textit{approximate} estimate by considering mentions affected by
homonyms if their gold KB entity has an associated name (a) which is a homonym and (b) whose surface form is highly
similar to the mention's one. Concretely, we consider a mention to be highly similar to a KB name if their normalized
Levenshtein distance \cite{Marzal1993ComputationON} is one (see Appendix \ref{sec:asm} for details).

\begin{table}
	\centering
	\resizebox{\columnwidth}{!}{
		\begin{tabular}{l|r}
			\toprule
			\multicolumn{2}{c}{\textbf{BELB: Biomedical Entity Linking Benchmark} \cite{BelbABiomediGarda2023}} \\
			\midrule
			\textsc{KB} (\textsf{entity type})                 & Homonyms                                       \\
			\multicolumn{1}{r|}{\texttt{Corpus}}               & \multicolumn{1}{l}{Affected mentions}          \\
			\midrule
			\textsc{CTD Diseases} (\textsf{Disease})           & 0.39\%                                         \\
			\multicolumn{1}{r|}{\texttt{NCBI Disease}}         & \multicolumn{1}{l}{1.25\%  (12 / 960)}         \\
			\multicolumn{1}{r|}{\texttt{BC5CDR (D)}}           & \multicolumn{1}{l}{0.18\% (8 / 4,363)}         \\
			\midrule
			\textsc{CTD Chemicals} (\textsf{Chemical})         & $\ll$1\%                                       \\
			\multicolumn{1}{r|}{\texttt{BC5CDR (C)}}           & \multicolumn{1}{l}{0\% (0 / 5,334)}            \\
			\multicolumn{1}{r|}{\texttt{NLM-Chem}}             & \multicolumn{1}{l}{0\% (0 / 11,514)}           \\
			\midrule
			\textsc{Cellosaurus} (\textsf{Cell line})          & 3.21\%                                         \\
			\multicolumn{1}{r|}{\texttt{BioID}}                & \multicolumn{1}{l}{3.47\% (30 / 864)}          \\
			\midrule
			\textsc{NCBI Gene} (\textsf{Gene})                 & 53.61\%                                        \\
			\multicolumn{1}{r|}{\texttt{GNormPlus}}            & \multicolumn{1}{l}{68.84\% (2,218 / 3,222)}    \\
			\multicolumn{1}{r|}{\texttt{NLM-Gene}}             & \multicolumn{1}{l}{66.1\% (1,804 / 2,729)}     \\
			\midrule
			\textsc{NCBI Taxonomy} (\textsf{Species})          & 0.04\%                                         \\
			\multicolumn{1}{r|}{\texttt{Linnaeus}}             & \multicolumn{1}{l}{0\% (0 / 1,430)}            \\
			\multicolumn{1}{r|}{\texttt{S800}}                 & \multicolumn{1}{l}{0\% (0 /767)}               \\
			\midrule
			\textsc{UMLS}                                      & 2.07\%                                         \\
			\multicolumn{1}{r|}{\texttt{MedMentions} (st21pv)} & \multicolumn{1}{l}{26.41\% (10,602 / 40,143)}  \\
			\bottomrule
		\end{tabular}
	}
	\caption{Relative number of homonyms and approximate estimate of the links they account for in BELB (test set).}\label{tab:ambiguous_mentions}
\end{table}

In Table \ref{tab:ambiguous_mentions} we report this estimate for corpora in BELB \cite{BelbABiomediGarda2023}, a BEL
benchmark comprising \numcorpora commonly used BEL corpora linked to \numkbs KBs (see Appendix \ref{sec:belb} for
details). Though some corpora have no such cases, we see that despite being only $\sim$2\% of \textsc{UMLS} names (see
Table \ref{tab:disamb_kbs} for exact counts), homonyms may account for up to 26\% of the links in the widely used
\texttt{MedMentions}, imposing an upper bound to the performance of name-based methods. The amount rises steeply to
more than 60\% for genes, for which homonyms are a well known aspect \cite{CrossSpeciesGWeiC2011} as the same gene can
be found in multiple species (see \S\ref{sec:method:disamb:xspecies}).

\section{Method}\label{sec:method}

We now introduce (i) our approach to replace KB homonyms with a disambiguated version and (ii) our architectural
enhancements to BioSyn, which together result in BELHD, our novel method for BEL.

\subsection{Homonym Disambiguation}\label{sec:method:disamb}

In BELHD we modify how the KB is represented to address the homonym issue. As mentioned, a homonym is a KB name $s$
s.t. $\vert \mathcal{V_{KB}}(s) \vert = n$ with $n>1$, where $n$ is the number of associated entities. For our approach
we draw inspiration from Wikipedia, where article titles that would otherwise be homonymous are disambiguated via
additional information in parentheses\footnote{\url{https://en.wikipedia.org/wiki/Wikipedia:Disambiguation}}.
Similarly, we expand every homonym into
into $n$ different versions, one for each entity, each augmented with a string having distinguishing information on the
entity it represents. For instance we replace \enquote{Discharge}
with \enquote{Discharge (Patient Discharge)} and \enquote{Discharge (Body Fluid Discharge)}. The augmentation strings
are other names in the KB. The procedure used to select them assumes that the KB reports for each entity, which of its
associated names is the \textit{preferred} one (usually the official name).


\begin{figure}
	\centering
	\includegraphics[scale=0.25]{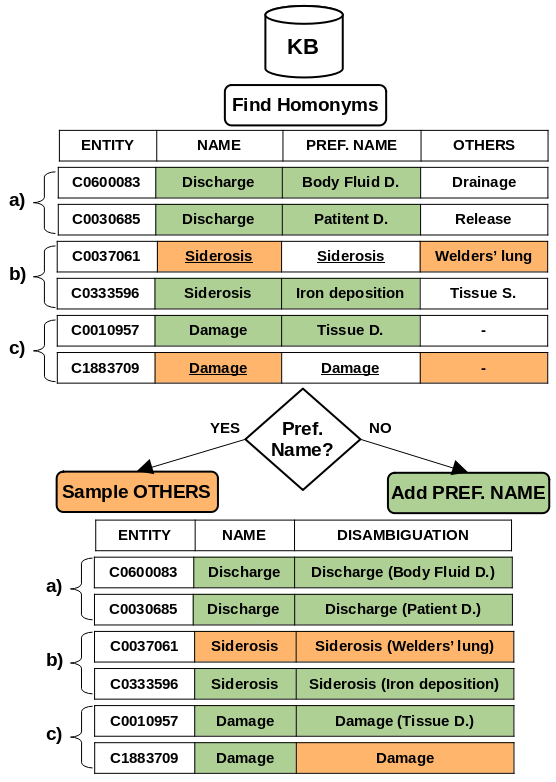}
	\caption{Illustration of our Homonym Disambiguation approach for biomedical KBs.}
	\label{fig:disamb_workflow}
\end{figure}

Our approach proceeds as follows (see Algorithm \ref{alg:disamb} for pseudocode). As shown in Figure
\ref{fig:disamb_workflow}, we first collect all names $s$ in the KB which are homonyms (see Appendix
\ref{sec:identify_homonyms} for details) and other names associated to the entities they refer to. If the homonym is
not the preferred name, we create disambiguated versions by adding the preferred name of the entities they represent,
as in (a). If instead the homonym is itself the preferred name, we select as disambiguation string the
\textit{shortest} name (which is not the homonym) associated to the entity, as in (b). This simple strategy can be
replaced with a KB-specific solution if metadata is available, e.g. selecting the name's official long form. Finally,
there are cases in which extending the name is not possible. This happens for homonyms which are preferred names but
not do provide additional names, as in (c). However, if a homonym has $n$ associated entities, we only need to create
$n-1$ disambiguated versions, with the unmodified one acting as the default meaning.

\subsubsection{Cross-species homonyms}\label{sec:method:disamb:xspecies}

The approach described above is valid for any biomedical KB whose entities specify a preferred name. However, in BEL
there exists a special class of homonyms which requires an additional step for complete disambiguation. Specific to
\textsf{Gene} and \textsf{Cell line}, these are cross-species homonyms. For instance, in \textsc{NCBI Gene}
\enquote{$\alpha$2microglobulin} can refer both to the human and cattle gene. As both genes have \enquote{A2M} as
preferred name, our approach would generate two still identical entries \enquote{$\alpha$2microglobulin
	(A2M)}\footnote{Though in principle this can happen in any KB, i.e. both name and preferred name are homonyms, in
	practice we find that this is specific to cross-species homonyms (see \S\ref{sec:results:hd}).}.

Cross-species homonyms play such a crucial role for these entity types that the corresponding KBs (\textsc{NCBI Gene}
and \textsc{Cellosaurus}) always report for each entity the species as well, in form of \textsc{NCBI Taxonomy}
entities. E.g. \textsc{NCBI Gene} for the human \enquote{A2M} reports 9606 (\enquote{human}) while for the cattle one
9913 (\enquote{cattle}). Therefore, before applying our first procedure (resolving intra-species homonyms), we identify
all cross-species homonyms (see Appendix \ref{sec:identify_homonyms} for details) and generate disambiguated versions
with the species name from \textsc{NCBI Taxonomy}. If a name is both an intra- and cross-species homonym, it will have
both disambiguation strings. E.g. \enquote{A2M} is also a secondary name for the human gene \enquote{IGHA2}, so our
approach will generate both \enquote{A2M ($\alpha$2microglobulin, human)} and \enquote{A2M (IGHA2, human)}.

\subsection{BELHD}\label{sec:method:belhd}

Here we first review the model upon which our approach is built, i.e. BioSyn, whose key aspect is its objective
function. \citet{BiomedicalEntiSung2020} observe that having a name-based search space entails that there are
potentially \textit{multiple} valid candidates for $m$. Therefore, they propose to train BioSyn as follows. For a given
mention $m$, BioSyn uses dense retrieval (maximum inner product search) to fetch a set of candidates name $C = \{c_{i},
	\cdots c_{n}\}$ from the (pre-encoded) KB\footnote{For simplicity, with a slight abuse of notation, we use $m$ and
	$c_{i}$ to refer both to the strings and their embeddings.}. The probability of each $c_{i}$ to be a correct link for
$m$ is defined as:
\begin{equation}
	\textrm{P}(c_{i} \vert m) =  \frac{
	\textrm{exp}({sim(m, c_{i})})
	}{
	\sum_{j=1}^{\vert C \vert} \textrm{exp}({sim(m, c_{j}))}
	}
\end{equation}
where $\vert C \vert$ is the size of $C$ and $sim$ is the inner product $\langle\cdot,\cdot\rangle$.
The model is optimized to minimize the following \textit{marginal maximum likelihood} (MML):
\begin{equation}
	l_{m} = -\textrm{log} \sum_{i=1}^{\vert C \vert}
	\mathbbm{1}_{[ \mathcal{V_{C}}(m) = \mathcal{V_{KB}}(c_{i}) ]}
	\textrm{P}(c_{i} \vert m)
\end{equation}
where  $\mathcal{V_{C}}: M \rightarrow E$ is a mapping returning the associated KB entity for a given mention $m \in M$,
while $\mathbbm{1}_{[ i = j ]} \in \{0,1\}$
is an indicator function evaluating to 1 iff $i=j$.
Intuitively speaking the objective function encourages the representations of $m$ and all candidates names to be close
in the dense space if they are associated the same KB entity.

\begin{figure}
	\centering
	\includegraphics[scale=0.25]{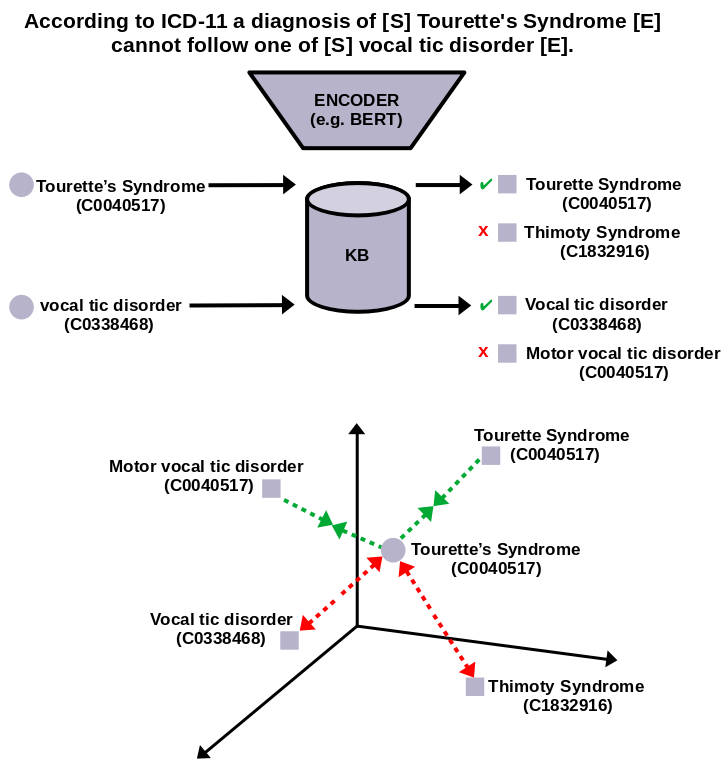}
	\caption{Overview of a BELHD training step with \textit{candidate sharing}.}
	\label{fig:belbert_training}
\end{figure}

To improve BioSyn, we keep the same training objective and introduce the following changes (see
\S\ref{sec:results:ablations} for ablation study).

\noindent \textbf{Retrieval} For simplicity, we described the BioSyn variant
using only dense retrieval. The original model combines both sparse (bi-gram TF-IDF) and dense
retrieval to fetch candidates from the KB. For BELHD, we rely exclusively on the latter
as the impact of sparse candidates in BioSyn is minimal.

\noindent \textbf{Context} In contrast to BioSyn, BELHD leverages contextual information.
For this we mark mentions boundaries with two special tokens: $[S]$ and $[E]$.
To obtain a single embedding we use mean pooling over $[S]$ and $[E]$.

\noindent \textbf{Projection head} To reduce memory footprint and increase search speed,
we introduce a projection head parametrized by a weight matrix $W \in \mathbb{R}^{h \times p}$
applied to \textit{both candidates and mentions} embeddings,
where $h$ and $p$ are the original and the projection head size, respectively.

\noindent \textbf{Candidate sharing} As reported by \citet{LearningDenseGillic2019}, hard negative mining
is crucial to successfully training (B)EL approaches based on contrastive learning (like BioSyn).
This entails retrieving difficult candidates for a mention from the KB at fixed intervals (e.g. at the end of each epoch).
We note however that, as co-occurring mentions are often related \cite{ALightweightNChen2021},
an important and easily exploitable training signal are candidates retrieved for other mentions in the same document.
For instance in Figure \ref{fig:belbert_training}, we see that by sharing candidates across mentions it is possible to explicitly
increase the similarity between \enquote{Tourette's Syndrome}
and \enquote{Motor vocal tic disorder}, which is another valid name for Tourette's,
but unlikely to be retrieved during hard negative mining due to its completely different surface form
(even for models not based on lexical matching).


Therefore, similar to \citet{HighlyParallelDeCao2021}, we encode the entire input text with all of its mentions.
However, while they use a Longformer \cite{LongformerTheBeltag2020} to support long sequences, we keep a BERT-based
architecture\footnote{To the best of our knowledge there is no Longformer pre-trained on biomedical data. The clinical
	version provided by \citet{AComparativeSLiYi2023} performed poorly on preliminary experiments.}. To overcome the
maximum sequence length we split each text into sentences and treat them as a single mini-batch. Then for each mention
$m_{i} \in M = \{m_{i} \cdots m_{n}\}$ we create pool of candidates $C_{i}$.
Half of the candidates are retrieved directly from the KB (as in BioSyn). The other half are instead selected from
$\bigcup_{\substack{j=1 \\ j \neq i}}^{\vert M \vert} C_{j}$ (candidates of other mentions different from those of
$m_{i}$), choosing the ones most similar to $m_{i}$ (measured by $sim$). Intuitively, if $\mathbbm{1}_{[
			\mathcal{V_{C}}(m_{i}) = \mathcal{V_{KB}}(c_{j}) ]} = 1$, these are hard positives (as the example in Figure
\ref{fig:belbert_training}), hard negatives otherwise.

\section{Experiments}\label{sec:exp}

In this section we perform empirical studies to (i) validate the effectiveness of our HD procedure and (ii) compare our
novel method against the state-of-the-art.

\subsection{Setting}\label{sec:exp:setting}

\noindent \textbf{Corpora and KBs} In all experiments we rely on BELB \cite{BelbABiomediGarda2023}
to access corpora and KBs. As a standardized benchmark,
BELB removes confounding factors such as differences in preprocessing and KB versions.
BELB consists of \numcorpora corpora linked to \numkbs KBs covering five entity types,
which allows for an evaluation more comprehensive than the one reported in previous studies.
In Table \ref{tab:ambiguous_mentions} we report BELB corpora and KBs used in our experiments (see Appendix
\ref{sec:belb} for details). As shown in \citet{BelbABiomediGarda2023}, current implementations of BEL systems cannot
scale to large KBs like \textsc{NCBI Gene} ($>$40M/100M entities/names).
Therefore, for \textsc{NCBI Gene}, we use the subsets determined by the species of the genes in \texttt{GNormPlus} and
\texttt{NLM-Gene} (see Appendix \ref{sec:kb_disamb_extra}). This reflects a common real-world use case, since often
only a specific subset of species is relevant for linking (e.g. human and mouse).

\noindent \textbf{Training} Unless otherwise stated, we ensure that
all models receive the same training signal by
(i) training them from scratch on BELB corpora and
(ii) avoiding corpus- or KB-specific pre-training.
For all baselines we rely on the original implementation provided by the authors.
Details w.r.t. models and training can be found in Appendix \ref{sec:models}.

\noindent \textbf{Metric} We report mention-level micro-average recall@1.
If for a given mention a method returns multiple entities as prediction,
instead of resorting to random sampling as proposed by \citet{zhang2022knowledge},
we consider the prediction incorrect.
This is because a primary use case of (B)EL is its deployment into pipelines \cite{BeedsLargeScWang2022},
where multiple entities cannot be used.

\subsection{Homonym Disambiguation}\label{sec:results:hd}

\begin{table*}
	\centering
	\resizebox{15cm}{!}{
		\begin{tabular}{l|l|l|l|l|l|l}
			\toprule
			             & \multicolumn{1}{c|}{\textsc{CTD Diseases}} & \multicolumn{1}{c|}{\textsc{CTD Chemicals}} & \multicolumn{1}{c|}{\textsc{Cellosaurus}} & \multicolumn{1}{c|}{\textsc{NCBI Gene}} & \multicolumn{1}{c|}{\textsc{NCBI Taxonomy}} & \multicolumn{1}{c}{\textsc{UMLS}} \\
			             & \multicolumn{1}{c|}{(\textsf{Disease})}    & \multicolumn{1}{c|}{(\textsf{Chemical})}    & \multicolumn{1}{c|}{(\textsf{Cell line})} & \multicolumn{1}{c|}{(\textsf{Gene})}    & \multicolumn{1}{c|}{(\textsf{Species})}     &                                   \\
			\midrule
			Entities     & 13,188                                     & 175,663                                     & 144,568                                   & 42,252,923                              & 2,491,364                                   & 3,464,809                         \\ \midrule Names & 88,548      & 451,410   &
			251,747      & 105,570,090                                & 3,783,882                                   & 7,938,833                                                                                                                                                             \\ Homonyms  & 349 (0.39\%)   & 2 ($\ll$1\%)                                                                                                                                                                                                                           & 8,070 (3.21\%) & 56,597,279
			(53.61\%)    & 1,422 (0.04\%)                             & 164,154 (2.07\%)                                                                                                                                                                                                    \\ \quad - pref. name & 1 & - & 502 & 822,103 & 27 & 12,530 \\ \quad -
			other        & 348                                        & 2                                           & 2,582                                     & 20,965,362                              & 1,395                                       & 151,624                           \\ \quad - cross-species & - & - & 5,416 & 56,265,683 & - & - \\
			\midrule
			Success rate & 100\%                                      & 100\%                                       & 100\%                                     & $\gg$99\% (1054)                        & 100\%                                       & $\gg$99\% (39)                    \\
			\bottomrule
		\end{tabular}
	}
	\caption{Number of names and homonyms (\% relative to names) categorized by cases (see \S\ref{sec:method:disamb}) in BELB KBs.
		The success rate is the ratio between homonyms which after augmentation are no longer homonyms and the original amount.
		Number of failures, i.e. names that are still homonyms (duplicates) is reported in parenthesis.
	}\label{tab:disamb_kbs}
\end{table*}

The effectiveness of our approach in disambiguating homonyms is measured by its success rate, i.e. the ratio between
homonyms which after expansion are no longer homonyms and the original ones in the KB. From Table \ref{tab:disamb_kbs}
we see that our approach can disambiguate virtually all homonyms in all \numkbs KBs (see Appendix
\ref{sec:kb_disamb_extra} for \textsc{NCBI Gene} subsets).
The failure cases are KB names having the same surface form \textit{after} our approach extends them with
distinguishing information,
and are not to be confused with the case where a preferred name is a homonym but has no alternative names, as case (c)
in \S\ref{sec:method:disamb}.
We find that these are caused by a combination of two factors. The first is related to the quality of the KBs.
\textsc{UMLS} and \textsc{NCBI Gene} are meta-KBs, i.e. they integrate entities from multiple KBs. In few cases, this
causes them to have two distinct entities having however little to no difference in terms of names. The second is using
the shortest alternative name as disambiguation string when a homonym is also the entity's preferred name. E.g. in
\textsc{UMLS} C0003663 and C0020316 have almost identical list of associated names. C0003663's preferred name is
\enquote{Aquacobalamin}, having as secondary name \enquote{Hydroxocobalamin}. However, \enquote{Hydroxocobalamin} is
also C0020316's preferred name, whose shortest alternative name is \enquote{Aquacobalamin}, giving raise to two
\enquote{Hydroxocobalamin (Aquacobalamin)}.



\subsubsection{Results}
\begin{table*}[!h]
	\centering
	\resizebox{\textwidth}{!}{
		\begin{tabular}{l|ll|ll|l|ll|ll|l}
			\toprule
			                                           & \multicolumn{2}{c|}{\textsc{CTD Diseases}} & \multicolumn{2}{c|}{\textsc{CTD Chemicals}} & \multicolumn{1}{c|}{\textsc{Cellosaurus}} & \multicolumn{2}{c|}{\textsc{NCBI Gene}} & \multicolumn{2}{c|}{\textsc{NCBI Taxonomy}} & \multicolumn{1}{c}{\textsc{UMLS}}                                                                                                          \\
			                                           & \multicolumn{2}{c|}{(\textsf{Disease})}    & \multicolumn{2}{c|}{(\textsf{Chemical})}    & \multicolumn{1}{c|}{\textsf{(Cell line)}} & \multicolumn{2}{c|}{(\textsf{Gene})}    & \multicolumn{2}{c|}{(\textsf{Species})}     &                                                                                                                                            \\
			\midrule
			                                           & \texttt{NCBI Disease}                      & \texttt{BC5CDR (D)}                         & \texttt{BC5CDR (C)}                       & \texttt{NLM-Chem}                       & \texttt{BioID}                              & \texttt{GNormPlus}                & \texttt{NLM-Gene}         & \texttt{S800}            & \texttt{Linnaeus}        & \texttt{MedMentions} \\
			\midrule
			\textit{Entity-based}                      &                                            &                                             &                                           &                                         &                                             &                                   &                           &                          &                          &                      \\
			\quad \textbf{arboEL} $\dagger$            & 80.00                                      & 84.87                                       & 87.40                                     & 71.76                                   & 95.02                                       & 34.64                             & 29.96                     & 78.62                    & 74.97                    & \underline{68.67}    \\
			\quad \textbf{KRISSBERT} $\dagger$ $\ddag$ & 82.80                                      & 85.0                                        & \textbf{95.10}                            & -                                       & -                                           & -                                 & -                         & -                        & -                        & 61.30                \\
			\midrule
			\textit{Name-based}                        &                                            &                                             &                                           &                                         &                                             &                                   &                           &                          &                          &                      \\
			\quad \textbf{BioSyn}                      & 79.90                                      & 84.83                                       & 84.57                                     & 70.35                                   & 80.79                                       & OOM                               & OOM                       & 82.79                    & 88.60                    & OOM                  \\
			\qquad + HD                                & 79.90                                      & $84.23_{-0.60}$                             & $85.00_{+0.43}$                           & $71.31_{+0.96}$                         & $81.60_{+0.9}$                              & OOM                               & OOM                       & $81.23_{-1.56}$          & $\mathbf{88.81}_{+0.21}$ & OOM                  \\
			\quad \textbf{GenBioEL}                    & 82.71                                      & \underline{88.29}                           & \underline{94.60}                         & \underline{75.00}                       & 94.79                                       & 6.80                              & 2.89                      & 88.27                    & 76.92                    & 41.16                \\
			\qquad + HD                                & $\underline{83.02}_{+0.31}$                & $88.20_{-0.09}$                             & $94.15_{-0.45}$                           & $74.10_{-0.90}$                         & $\underline{96.30}_{+1.51}$                 & $\underline{66.08}_{+59.28}$      & $\mathbf{66.43}_{+63.54}$ & $\mathbf{89.96}_{+1.69}$ & $77.62_{-0.30}$          & $64.59_{+23.43}$     \\
			\quad \textbf{BELHD (ours)}                & \textbf{87.60}                             & \textbf{89.23}                              & 92.93                                     & \textbf{82.39}                          & \textbf{96.99}                              & \textbf{77.84}                    & \underline{59.03}         & \underline{84.35}        & \underline{81.89}        & \textbf{70.58}       \\
			\bottomrule
		\end{tabular}
	}
	\caption{Performance of all models on BELB corpora (test set).
		\textbf{Bold} and \underline{underlined} indicate best and second best score, respectively.
		HD: Homonym Disambiguation (\S\ref{sec:method:disamb}). OOM: out-of-memory ($>$200GB) $\dagger$ Without cross-encoder
		reranking $\ddag$ Authors provide code only for the \enquote{supervised} variant\protect\footnotemark, i.e. entity prototypes
		are based on train/development mentions. As this variant cannot link zero-shot entities we report results from Table 7
		in \cite{zhang2022knowledge} for a fair comparison.
	}~\label{tab:results:mention}
\end{table*}

From Table \ref{tab:results:mention} we see that BELHD is the overall best approach, outperforming all baselines in
\numoutperform out of \numcorpora corpora, with GenBioEL as second. Our results are in line with general-domain entity
linking, where name-based approaches generally outperform those relying on entity representations
\cite{OnTheSurprisiSchuma2023}.
Importantly, we note that our HD solution can be used with any name-based model. Notably, when equipped with HD,
GenBioEL's performance increases by 63.54pp on the homonym-rich \texttt{NLM-Gene}, outperforming all other methods,
including BELHD. This can be attributed to the fact that the corpus was specifically created to test models on
cross-species homonyms \cite{Islamaj2021}. Key to success on the task is contextual information, primarily in form of
species mentions. While GenBioEL processes the entire text, BELHD uses sentences (see \S\ref{sec:method:belhd}), thus
being unable to access species information if it does not occur in the same sentence as the gene mention.
Finally, we see that HD as minimal to no effect in BioSyn, since it does not use context. BioSyn is however the best
model for the \texttt{Linnaues} corpus. This can be explained by the fact that \texttt{Linnaeus} was created
specifically for the development of dictionary-based approaches \cite{S1000ABetterLuoma2023}, giving a strong advantage
to methods using string-matching like BioSyn.

\subsubsection{Ablation study}\label{sec:results:ablations}

\footnotetext{\url{https://huggingface.co/microsoft/BiomedNLP-KRISSBERT-PubMed-UMLS-EL}}

\begin{table}[!h]
	\centering
	\resizebox{5cm}{!}{
		\begin{tabular}{l|l}
			\toprule
			                          & \multicolumn{1}{c}{\textsc{NCBI Gene}} \\
			                          & \multicolumn{1}{c}{(\textsf{Gene})}    \\
			\midrule
			                          & \multicolumn{1}{c}{\texttt{NLM-Gene}}  \\
			\midrule
			\textbf{BELHD (ours)}     & 59.03                                  \\
			\quad - HD                & $6.67_{-52.8}$                         \\
			\quad - context           & $32.72_{-26.31}$                       \\
			\quad - candidate sharing & $56.83_{-2.20}$                        \\
			\quad - projection head   & $58.74_{-0.29}$                        \\
			\bottomrule
		\end{tabular}
	}
	\caption{Ablation study of improvements over BioSyn introduced in BELHD (see \S\ref{sec:method:belhd}).}\label{tab:belbert_ablations}
\end{table}

Table \ref{tab:belbert_ablations} reports our ablation study of BELHD improvements over BioSyn (see
\S\ref{sec:method:belhd}). We see that the most important component is HD, which is critical for a homonym-rich entity
type like \textsf{Gene}, while the use of contextual information is second. The third strongest improvement is brought
by \textit{candidate sharing}. This confirms that leveraging the relatedness of co-occurring mentions enhances training
signal improving overall results. Finally, the projection head, despite reducing the embedding dimensionality, does not
hurt performance.

\subsubsection{Ad-hoc solutions for Homonym Disambiguation}

\begin{table}[!htbp]
	\centering
	\resizebox{\columnwidth}{!}{
		\begin{tabular}{l|l|ll}
			\toprule
			                          & \multicolumn{1}{c|}{\textsc{UMLS}} & \multicolumn{2}{c}{\textsc{NCBI Gene}}                             \\
			                          &                                    & \multicolumn{2}{c}{(\textsf{Gene})}                                \\
			\midrule
			                          & \texttt{MedMentions}               & \texttt{GNormPlus}                     & \texttt{NLM-Gene}         \\
			\midrule
			\textbf{arboEL}           & 68.67                              & 34.64                                  & 29.96                     \\
			\quad + AR                & $68.85_{+0.18}$                    & $35.66_{+1.02}$                        & $30.30_{+0.34}$           \\
			\quad + Species $\dagger$ & -                                  & $\mathbf{47.83_{+13.19}}$              & $\mathbf{40.82_{+10.86}}$ \\
			\midrule
			\textbf{GenBioEL}         & 41.16                              & 6.80                                   & 2.89                      \\
			\quad + AR                & $42.01_{+0.85}$                    & $7.67_{+0.87}$                         & $3.48_{+0.59}$            \\
			\quad + SA                & -                                  & $65.70_{+58.90}$                       & $61.78_{+58.89}$          \\
			\quad + HD                & $\mathbf{64.59_{+23.43}}$          & $\mathbf{66.08_{+59.28}}$              & $\mathbf{66.43_{+63.54}}$ \\
			\midrule
			\textbf{BELHD w/o HD}     & 57.23                              & 13.90                                  & 6.67                      \\
			\quad + AR                & $59.50_{+2.27}$                    & $15.98_{+2.08}$                        & $6.85_{+0.18}$            \\
			\quad + SA                & -                                  & $43.67_{+29.77}$                       & $42.58_{+35.91}$          \\
			\quad + HD                & $\mathbf{70.58_{+13.35}}$          & $\mathbf{77.84_{+63.94}}$              & $\mathbf{59.03_{+52.35}}$ \\
			\bottomrule
		\end{tabular}
	}
	\caption{Effect of different strategies to handle homonyms.
		HD: Homonym Disambiguation (ours),
		AR: Abbreviation Resolution \cite{AbbreviationDeSohn2008},
		SA: Species Assignment \cite{AssigningSpeciLuoL2022}.
		$\dagger$ Include species name into entity representation \cite{AComprehensiveKartch2023}
	}\label{tab:disamb_vs_adhoc}
\end{table}


Here we compare ad-hoc methods to handle homonyms (see Section \ref{sec:rw}) with our general HD approach on the
corpora most affected by homonyms. We perform abbreviation resolution (AR) with Ab3P (replacing abbreviations with
their long form) and retrain all models. Secondly, we identify and assign species to gene mentions with
SpeciesAssignment (SA) and filter predictions accordingly. As AR resolves difficult mentions which are not necessarily
affected by homonyms (see Section \ref{sec:homonyms}) we include the entity-based arboEL in the comparison. As SA is
specific to name-based methods, for arboEL we include species name in entity representations as proposed by
\cite{AComprehensiveKartch2023}.
From Table \ref{tab:disamb_vs_adhoc} we see that HD delivers the best results across corpora. It significantly
outperforms AR, confirming that addressing a wider range of homonyms is critical. Interestingly, HD outperforms the
highly specialized SA approach as well. We argue that this is due to species information not being always explicitly
expressed \cite{Sr4gnASpecieWeiC2012}, upon which SA relies. HD instead is more versatile, allowing the model to
\textit{learn} useful contextual patterns beyond explicit species mentions.

\subsubsection{Entity- vs name-based dense retrieval}

\begin{table}[!htbp]
	\centering
	\resizebox{5cm}{!}{
		\begin{tabular}{l|c}
			\toprule
			                            & \textsc{CTD Diseases} \\
			                            & (\textsf{Disease})    \\
			\midrule
			                            & \texttt{NCBI Disease} \\
			\midrule
			\textbf{Entity-based}       & 72.19                 \\
			\textbf{Name-based} (w/ HD) & 78.85                 \\
			\bottomrule
		\end{tabular}
	}
	\caption{Difference between entity-based and name-based (with HD) approach with same experimental conditions (input representation, model, candidate selection).
	}\label{tab:name_vs_entity_retrieval}
\end{table}

Large corpora like \texttt{MedMentions} are the exception in BEL (see Appendix \ref{sec:belb}).
We therefore aim to compare the name-based and entity-based approach when limited training data is available. To
exclude potential confounding factors (input representation, model type, candidate selection) we follow the
experimental setting of \citet{ScalableZeroSWuLe2020} and train two identical bi-encoders\footnote{We focus on dense
	retrieval since \citet{decao2021autoregressive} already shown their generative approach performs poorly if trained to
	generate unique IDs instead of names.} on the same input (see Appendix \ref{sec:name_vs_entity_retrieval} for details).
The models differ only in the KB representation and, consequently, in the objective function: one uses standard cross
entropy (entity-based) while the other MML (name-based: see \S\ref{sec:method:belhd}). Results in Table
\ref{tab:name_vs_entity_retrieval} suggest that name-based are inherently more sample-efficient, as they can directly
leverage surface similarities between mentions and KB names, while entity-based require more training data to optimize
entity representations.

\section{Discussion}\label{sec:discussion}

We introduce BELHD, a novel method for biomedical entity linking.
Our experiments show that BELHD outperforms all baselines in \numoutperform out of \numcorpora BEL corpora.
We stress that we retrain all models on BELB with the code and hyperparameters provided by the authors
and thus numbers we report may differ from those found in the original publications (see Section \ref{sec:limitations}).
However, we believe that this setting is the best approximation  
to fairly compare across methods since, as reported by \citet{BelbABiomediGarda2023},
BEL studies present stark differences in preprocessing and experimental setups
making comparison based on published numbers problematic.

Secondly, we note that our study focuses on \textit{first-stage} BEL, i.e. candidate generator methods.
Reranking, e.g. with a cross-encoder \cite{ScalableZeroSWuLe2020},
is a further enhancement \textit{orthogonal} to the choice of the generator. 
Thus using a cross-encoder only for one generator (arboEL) as in \cite{AComprehensiveKartch2023} gives it an unfair advantage
producing biased results. Either different rerankers
are compared with the same set of candidates or the same reranker assesses the quality of different sets.
We leave an analysis of the latter as future work.



\section{Conclusion}\label{sec:conclusion}

We highlight how homonyms in biomedical KBs
significantly impact performance of BEL methods
returning KB names as predictions.
We introduce BELHD, a novel BEL approach based on BioSyn \cite{BiomedicalEntiSung2020} 
outperforming all baselines in \numoutperform out of \numcorpora corpora.
We show that its primary feature HD is a general solution improving results in other name-based methods as well.


%

\section{Limitations}\label{sec:limitations}

A limitation of our work is the assumption on the biomedical KBs required
by the HD procedure. That is, the KB must specify which is the preferred name of a given entity.
Though, to the best of our knowledge, virtually all biomedical KBs meet this assumption,
the HD procedure is therefore not strictly KB-agnostic.
Secondly,  due to quality issues in biomedical KBs,
i.e. entities having almost identical lists of associated names,
it is not possible to remove all homonyms (see \S\ref{sec:method:disamb}).
This could be mitigated by using more sophisticated strategies to select the disambiguation string, 
e.g. using the least similar name determined by Levenshtein distance, which we leave as future work .

As already mentioned in Section \ref{sec:discussion},
an important limitation of our study is the lack of hyperparameter exploration of all baselines.
Due to the high computational resources necessary to train BEL models
we are limited to rely on the default ones reported by the authors.
It is therefore possible that optimizing them may result in better numbers.
Additionally, due to the large amount of combinations of corpora, models and ad-hoc components
we are limited to our best effort and do not report results with multiple seeds.
We note as well that we train all models from scratch on BELB corpora 
avoiding corpus- or KB-specific pre-trained weights. This is because, as noted by \citet{milich-akbik-2023-zelda},
different methods use different pre-training strategies and different data, ultimately impairing direct comparison.
It is therefore possible that GenBioEL, with its \textit{KB-Guided Pre-training},
and BioSyn, by using SapBERT weights \cite{SelfAlignmentLiuF2021}, may achieve higher results.
Finally, we note that there exists entity-specific BEL models, e.g. GNorm2 \cite{Gnorm2AnImprWeiC2023} for genes.
We do not evaluate them as the primary focus of this study are entity-agnostic BEL models.

\section{Ethical Considerations}

A primary use case of biomedical entity linking is its deployment in information extraction pipelines, 
which in turn are deployed in application to facilitate the navigation of the scientific literature by biomedical researchers.
As shown by an existing large body of work, language models (such as the one used in our work) may present biases.
Therefore, a potential harmful downstream consequence may be casused by systems' errors caused by those biases.
For instance, a system incorrectly linking a mention of \enquote{Postmenopausal Osteoporosis} (C0029458)
to the general \enquote{Osteoporis} (C0029456) due to an implicit gender bias 
may prevent a relevant publication for the condition to be found by researchers.
Secondly, if results of extraction pipelines are used to populate biomedical knowledge bases,
and in turn these resources are used to train other models, 
these implicit biases may be further propogated and amplified.

\section*{Acknowledgements}

Samuele Garda is supported by the \textit{Deutsche Forschungsgemeinschaft} as part of the
research unit \enquote{Beyond the Exome}.
We would like to thank Mario S\"anger and Leon Weber-Genzel for useful feedback on the manuscript.

\bibliography{reference}
\appendix

\section{Approximate String Matching}\label{sec:asm}

We first preprocess mentions and KB names by
lowercasing and removing all non alphanumeric characters.
We compute a score in $[0,1]$ (the higher the more similar)
between each mention and the names associated with its gold KB entity
and select the KB names with a similarity score of 1. 
The score is the defined as $\frac{D}{\vert s_{i} \vert  + \vert s_{j} \vert }$,
where $D$ is the Levenshtein distance (with a substitution weight of 2)
between strings $s_{i}$ and $s_{j}$ and $\vert s_{i} \vert$ is the number of characters in $s_{i}$.
We use the implementation provided by \url{https://github.com/maxbachmann/RapidFuzz}.

\section{Biomedical Entity Linking Benchmark}\label{sec:belb}

\begin{table*}[!htbp]
	\centering
	\resizebox{10cm}{!}{
		\begin{tabular}{l|c|c|c}
			\toprule

			\textsf{Entity type}                                        & \multicolumn{3}{c}{}                                       \\
			\quad \textsc{KB}                                           & Entities             & Names       & Avg. names per entity \\
			  \midrule
			\textsf{Disease}                                            &                      &             &                       \\
			\quad \textsc{CTD Diseases} \cite{ComparativeToxDavis2023}  & 13,188               & 88,548      & 6.71                  \\
			  \midrule                                                                                                             %
			\textsf{Chemical}                                           &                      &             &                       \\
			\quad \textsc{CTD Chemicals} \cite{ComparativeToxDavis2023} & 175,663              & 451,410     & 2.56                  \\
			  \midrule                                                                                                             %
			\textsf{Cell line}                                          &                      &             &                       \\
			\quad \textsc{Cellosaurus} \cite{TheCellosaurusBairoc2018}  & 144,568              & 251,747     & 1.74                  \\
			  \midrule                                                                                                             %
			\textsf{Species}                                            &                      &             &                       \\
			\quad \textsc{NCBI Taxonomy} \cite{TheNcbiTaxonoScott2012}  & 2,491,364            & 3,783,882   & 1.51                  \\
			  \midrule                                                                                                             %
			\textsf{Gene}                                               &                      &             &                       \\
			\quad \textsc{NCBI Gene} \cite{Brown2015}                   & 42,252,923           & 105,570,090 & 2.49                  \\
			\qquad \texttt{GNormPlus} subset                            & 703,858              & 2,455,772   & 3.48                  \\
			\qquad \texttt{NLM-Gene} subset                             & 873,015              & 2,913,456   & 3.33                  \\
			  \midrule                                                                                                             %
			\textsc{UMLS}                                               &                      &             &                       \\
			\quad \textsc{UMLS} \cite{TheUnifiedMedBodenr2004}          & 3,464,809            & 7,938,833   & 2.29                  \\
			  \bottomrule
		\end{tabular}
	}
	\caption{Overview of the KBs available in BELB according to their entity type.
		We report the number of entities, names and average name per entities}~\label{tab:belb_kbs}
\end{table*}

\begin{table*}[!h]
	\centering
	\resizebox{15cm}{!}{
		\begin{tabular}{l|c|c|c}
			\toprule
			\textsf{Entity type}                                            & \multicolumn{3}{c}{}                                                             \\
			\quad \texttt{Corpus}                                           & Documents (train / dev / test) & Mentions (train / dev / test) & 0-shot mentions \\
			\midrule
			\textsf{Disease}                                                &                                &                                                 \\
			\quad \texttt{NCBI Disease} \cite{Dogan2014}                    & 592 / 100 / 100                & 5,133 / 787 / 960             & 150 (15.62\%)   \\
			\quad \texttt{BC5CDR} (D) \cite{Li2016a}                        & 500 / 500 / 500                & 4,149 / 4,228 / 4,363         & 388 (8.89\%)    \\
			\midrule
			\textsf{Chemical}                                               &                                &                                                 \\
			\quad \texttt{BC5CDR} (C)\cite{Li2016a}                         & 500 / 500 / 500                & 5,148 / 5,298 / 5,334         & 1,038 (19.46\%) \\
			\quad \texttt{NLM-Chem} $\dagger$ \cite{NlmChemBc7MIslama2022}  & 80 / 20 / 50                   & 20,796 / 5,234 / 11,514       & 3,908 (33.94\%) \\
			\midrule
			\textsf{Cell line}                                              &                                &                                                 \\
			\quad \texttt{BioID} $\ddag$ \cite{arighi2017bio}               & 231 / 59 / 60                  & 3,815 / 1,096 / 864           & 158 (18.29\%)   \\
			\midrule
			\textsf{Species}                                                &                                &                                                 \\
			\quad \texttt{Linnaeus} $\dagger$ \cite{LinnaeusASpeGerner2010} & 47 / 17 / 31                   & 2,115 / 705 / 1,430           & 385 (26.92\%)   \\
			\quad \texttt{S800} \citep{Pafilis2013}                         & 437 / 63 / 125                 & 2,557 / 384 / 767             & 363 (47.33\%)   \\
			\midrule
			\textsf{Gene}                                                   &                                &                                                 \\
			\quad \texttt{GNormPlus} \cite{Wei2015}                         & 279 / 137 / 254                & 3,015 / 1,203 / 3,222         & 2,822 (87.59\%) \\
			\quad \texttt{NLM-Gene} \cite{Islamaj2021}                      & 400 / 50 / 100                 & 11,263 / 1,371 / 2,729        & 1,215 (44.52\%) \\
			\midrule
			\textsc{UMLS}                                                   &                                &                                                 \\
			\quad \texttt{MedMentions} (st21pv) \citep{mohanmedmentions}    & 2,635 / 878 / 879              & 122,178 / 40,864 / 40,143     & 8,167 (20.34\%) \\
			\bottomrule
		\end{tabular}
	}
	\caption{Overview of the corpora available in BELB with number of documents, mentions and 0-shot mentions (mentions linked to an entity not in the train/development set).
		Pairing of corpora and KB is determined by the \textsf{entity type}.
		$\ddag$ Full text
		$\ddag$ Figure captions}~\label{tab:belb_corpora}
\end{table*}

BELB is a biomedical entity linking benchmark introduced by \citet{BelbABiomediGarda2023}. It provides access to 10
corpora linked to \numkbs knowledge bases and spanning five entity types: \textsf{Gene}, \textsf{Disease},
\textsf{Chemical}, \textsf{Species} and \textsf{Cell lines}: see Table \ref{tab:belb_kbs} and Table
\ref{tab:belb_corpora} for an overview of the KBs and corpora, respectively. The key feature of BELB is its
standardized preprocessing of corpora and KBs, offering tight integration between the two. This makes it a standardized
testbed which removes confounding factors such as differences in preprocessing and KB versions. All corpora consists of
biomedical publications in English. For a detailed description and license information of each corpus and KB we refer
the reader to the original publication.

\section{Identify homonyms}\label{sec:identify_homonyms}

\begin{listing}[!htbp]
	\begin{minted}{sql}
CREATE TABLE kb(
    uid INTEGER PRIMARY KEY,
    -- entity label
    identifier INTEGER NOT NULL, 
    -- pref. name (0), abbr. (1), ...
    description INTEGER NOT NULL, 
    name TEXT NOT NULL,
    -- NCBI Taxonomy entity
    species INTEGER DEFAULT NULL
)
\end{minted}
	\caption{Schema used in BELB to store biomedical KBs.}\label{lst:schema}
\end{listing}

In Listing \ref{lst:schema} we present the unified schema provided in BELB used to store all biomedical KB. Each name
is associated to an identifier (entity label) and a description, i.e. whether it is e.g. the preferred name or the
abbreviated form. In case of KBs having cross-species homonyms (see \S\ref{sec:method:disamb:xspecies}) BELB stores the
name of the associated species entity coming from \textsc{NCBI Taxonomy}.



\begin{listing}[!htbp]
	\begin{minted}{sql}
-- Homonyms
SELECT name FROM kb 
    GROUP BY name,species 
    HAVING count(*)>1

-- Cross-species homonyms
SELECT name FROM kb 
    GROUP BY name 
    HAVING count(*)>1 AND
    COUNT(DISTINCT(species))>1;
\end{minted}
	\caption{Example SQL queries to compute set of homonyms with BELB KBs.}\label{lst:identify_homonyms}
\end{listing}

Listing \ref{lst:identify_homonyms} shows how using BELB we generate the set of homonyms and cross-species homonyms. In
the group by of the first query we include \enquote{species} to ensure that the homonyms belong to the same species
(intra-species), e.g. \enquote{BRI3} can be either \textsc{NCBI Gene} 81618 or 25798, both however are \textit{human}
genes. The second query instead specifically identifies only cross-species homonyms, i.e. it constraints names to have
different associated species.

\section{Pseudocode for Homonym Disambiguation procedure}

\begin{algorithm}[!htbp]
	\caption{Pseudocode our disambiguation approach.}\label{alg:disamb}
	\begin{algorithmic}[1]
		\Require $\mathcal{H}$ \Comment{Pre-computed set of homonyms}
		\Require $\mathcal{E}$ \Comment{Entities}
		\ForEach {$e \in \mathcal{E} $}
		\State $\mathcal{S} \gets \textnormal{get\_names}(e)$
		\ForEach {$s \in \mathcal{S} $}
		\If{$s \in \mathcal{H}$}
		\State $p \gets \textnormal{get\_preferred\_name}(\mathcal{S})$
		\If{$s = p$}
		\LineComment{s.t. $s$ != $p$}
		\State $d \gets \textnormal{get\_longest}(s,\mathcal{S})$
		\Else
		\State $d \gets p$
		\EndIf
		\State $s \gets \textnormal{concatenate}(s,d)$
		\EndIf
		\EndFor
		\EndFor
	\end{algorithmic}
\end{algorithm}

In Algorithm \ref{alg:disamb} we present the pseudocode of our approach to resolve homonyms in biomedical KBs present
in \S\ref{sec:method:disamb}. The pseudocode for the cross-species procedure described in
\S\ref{sec:method:disamb:xspecies} can be easily derived from it.


\section{Models and training details}\label{sec:models}

Here we report training details for all models considered in our study. We stress that we retrain all models on BELB
with the code provided by the original authors (see Table \ref{tab:systems:overview} for links to implementations). All
experiments were performed on two NVIDIA A100 GPUs.

\begin{table}[!htbp]
	\centering
	\resizebox{\columnwidth}{!}{
		\begin{tabular}{l|l}

			\toprule
			               & Implementation (link)                                                                 \\
			\midrule
			\quad arboEL   & \href{https://github.com/dhdhagar/arboEL}{https://github.com/dhdhagar/arboEL}         \\
			\quad GenBioEl & \href{https://github.com/Yuanhy1997/GenBioEL}{https://github.com/Yuanhy1997/GenBioEL} \\
			\quad BioSyn   & \href{https://github.com/dmis-lab/BioSyn}{https://github.com/dmis-lab/BioSyn}         \\
			\bottomrule
		\end{tabular}
	}
	\caption{Implementation links of the biomedical entity linking models used in our experiments.}~\label{tab:systems:overview}
\end{table}

\noindent \textbf{BioSyn} uses the default hyper-parameters provided by the authors.
Unlike in the original study, we exclusively train on the train split of corpora (no development).
Total amount of parameters: 110M.

\noindent \textbf{GenBioEl} uses different values for learning rate and warmup steps for \texttt{NCBI Disease} and \texttt{BC5CDR}.
We cannot perform a full hyper-parameter search for each corpus,
and therefore select the values that work best for both corpora,
i.e. a learning rate of $1e-5$ and 500 warmup steps. Total amount of parameters: 400M.

\noindent \textbf{arboEL}'s inference procedure is parametrized by
the number of $k$ nearest neighbor used to construct the graph (determining which pairs of nodes are connected).
The implementation provided by the authors runs the inference trying different $k \in \{0, 1, 2, 4, 8\}$.
For fair comparison with other models, we do not perform any hyperparameter optimization
and hence report the score for $k=0$. Total amount of parameters: 110M.

\noindent \textbf{BELHD} keeps all BioSyn hyper-parameter besides (i)
the number of training epochs which we increase from 10 to 20 and the number of candidates $k$ for each mention $m_{i}$,
which we set to 32: 16 mention-specific and 16 from candidate sharing (see \S\ref{sec:method:belhd})
Like in Biosyn, the KB is re-encoded at the and of each training epoch to keep the name embeddings consistent with
the updated model parameters \cite{pmlr-v119-guu20a}.
The only exception is \texttt{MedMentions}, for which, due to its large size, we use 10 epochs and re-encode
the KB every 1000 steps. The dimensionality of the projection head is set to 128. In order to fit the entire text unit at once,
we split it into sentences (with segtok\footnote{\url{https://github.com/fnl/segtok}})
and treat it as a single mini-batch, using on gradient accumulation to achieve a larger batch size,
which we set to 8. We rely on FAISS \cite{johnson2019billion} for efficient exact maximum inner product search.
Total amount of parameters: 110M (projection head has 1e4 parameters).

\subsection{Name- vs entity-based dense retrieval}\label{sec:name_vs_entity_retrieval}

Following \citet{ScalableZeroSWuLe2020} both models take as input a mention centered in a fixed context window. The
input is truncated (left and/or right) to be of maximum 128 tokens. The input to the candidate encoder (entity or name)
is maximum 128 tokens. Each model uses in-batch negatives with an additional 10 hard negatives mined during training,
i.e the top 10 predicted entities/names for each training mention. Both models are trained with a mini batch size of 32
for 10 epochs.

\section{Corpus-specific NCBI Gene subsets}\label{sec:kb_disamb_extra}

\begin{table}[!htbp]
	\resizebox{\columnwidth}{!}{
		\begin{tabular}{l|ll}
			\toprule
			                      & \multicolumn{2}{c}{\textsc{NCBI Gene}}                       \\
			                      & \multicolumn{2}{c}{(\textsf{Gene})}                          \\
			\midrule
			                      & \texttt{GNormPlus}                     & \texttt{NLM-Gene}   \\
			\midrule
			Names                 & 2,455,772                              & 2,913,456           \\
			\midrule
			Homonyms              & 1,163,255 (47.37\%)                    & 1,479,719 (50.79\%) \\
			\quad - pref. name    & 33,919                                 & 35,032              \\
			\quad - other         & 323,824                                & 39,1290             \\
			\quad - cross-species & 1,094,531                              & 1,410,006           \\
			\midrule
			Success rate          & $>$99\% (520)                          & $>$99\%  (523)      \\
			\bottomrule
		\end{tabular}
	}
	\caption{Equivalent to Table \ref{tab:disamb_kbs} for \textsc{NCBI Gene} corpus-specific subsets.}\label{tab:disamb_kbs_gene_subset}
\end{table}

\begin{table}[!htbp]
	\resizebox{\columnwidth}{!}{
		\begin{tabular}{l|l|l}
			\toprule
			\multicolumn{2}{c|}{\textsc{NCBI Taxonomy}} &                                                                                  \\
			\midrule
			Entity                                      & Name                                      & Corpora                              \\
			\midrule
			3055                                        & Chlamydomonas reinhardtii                 & \texttt{NLM-Gene}                    \\
			3702                                        & thale cress                               & \texttt{GNormPlus},\texttt{NLM-Gene} \\
			3847                                        & soybean                                   & \texttt{GNormPlus}                   \\
			4896                                        & fission yeast                             & \texttt{GNormPlus},\texttt{NLM-Gene} \\
			6239                                        & Caenorhabditis elegans                    & \texttt{GNormPlus},\texttt{NLM-Gene} \\
			6956                                        & European house dust mite                  & \texttt{NLM-Gene}                    \\
			7227                                        & fruit fly <Drosophila melanogaster>       & \texttt{GNormPlus},\texttt{NLM-Gene} \\
			7955                                        & zebrafish                                 & \texttt{GNormPlus},\texttt{NLM-Gene} \\
			8355                                        & African clawed frog                       & \texttt{GNormPlus},\texttt{NLM-Gene} \\
			8364                                        & tropical clawed frog                      & \texttt{GNormPlus},\texttt{NLM-Gene} \\
			9031                                        & chicken                                   & \texttt{GNormPlus},\texttt{NLM-Gene} \\
			9606                                        & human                                     & \texttt{GNormPlus},\texttt{NLM-Gene} \\
			9615                                        & dog                                       & \texttt{NLM-Gene}                    \\
			9823                                        & pig                                       & \texttt{GNormPlus},\texttt{NLM-Gene} \\
			9913                                        & cattle                                    & \texttt{GNormPlus},\texttt{NLM-Gene} \\
			9940                                        & sheep                                     & \texttt{NLM-Gene}                    \\
			9986                                        & rabbit                                    & \texttt{GNormPlus},\texttt{NLM-Gene} \\
			10029                                       & Chinese hamster                           & \texttt{NLM-Gene}                    \\
			10089                                       & Ryukyu mouse                              & \texttt{NLM-Gene}                    \\
			10090                                       & house mouse                               & \texttt{GNormPlus},\texttt{NLM-Gene} \\
			10116                                       & Norway rat                                & \texttt{GNormPlus},\texttt{NLM-Gene} \\
			10298                                       & Herpes simplex virus type 1               & \texttt{GNormPlus}                   \\
			11676                                       & Human immunodeficiency virus 1            & \texttt{GNormPlus},\texttt{NLM-Gene} \\
			11709                                       & Human immunodeficiency virus 2            & \texttt{NLM-Gene}                    \\
			11908                                       & Human T-cell leukemia virus type I        & \texttt{GNormPlus}                   \\
			41856                                       & Hepatitis C virus genotype 1              & \texttt{GNormPlus}                   \\
			51031                                       & New World hookworm                        & \texttt{NLM-Gene}                    \\
			81972                                       & Arabidopsis lyrata subsp. lyrata          & \texttt{NLM-Gene}                    \\
			333760                                      & Human papillomavirus type 16              & \texttt{GNormPlus}                   \\
			511145                                      & Escherichia coli str. K-12 substr. MG1655 & \texttt{GNormPlus}                   \\
			559292                                      & Saccharomyces cerevisiae S288C            & \texttt{GNormPlus},\texttt{NLM-Gene} \\
			2886926                                     & Escherichia phage P1                      & \texttt{NLM-Gene}                    \\
			\bottomrule
		\end{tabular}
	}
	\caption{\textsc{NCBI Gene} subsets determined by the species (\textsc{NCBI Taxonomy} entities) of the gene mentions in \texttt{GNormPlus} and \texttt{NLM-Gene}}\label{tab:ncbi_gene_subsets}
\end{table}

For the \textsc{NCBI Gene} subsets determined by the species of the genes in \texttt{GNormPlus} and \texttt{NLM-Gene}
(see \S\ref{sec:exp:setting}) we report in Table \ref{tab:disamb_kbs_gene_subset} the number of homonyms and the
success rate of our disambiguation approach. In Table \ref{tab:ncbi_gene_subsets} we report the \textsc{NCBI Gene}
subsets determined by the species (\textsc{NCBI Taxonomy} entities) of the gene mentions in \texttt{GNormPlus} and
\texttt{NLM-Gene}.

\end{document}